\documentclass[runningheads]{llncs}
\usepackage{graphicx}
\usepackage{subfig}

\begin{document}

\title{Entity Extraction with Knowledge from Web Scale Corpora}
\author{Zeyi Wen\inst{1} \and
Zeyu Huang\inst{2} \and
Rui Zhang\inst{2}}
\authorrunning{Z. Wen, Z. Huang and R. Zhang}
\institute{$^1$The University of Western Australia, $^2$The University of Melbourne\\
\email{zeyi.wen@uwa.edu.au, \{z.huang56@student.,rui.zhang@\}unimelb.edu.au}}

\maketitle

\begin{abstract}
Entity extraction is an important task in text mining and natural language processing. A popular method for entity extraction is by comparing substrings from free text against a dictionary of entities. In this paper, we present several techniques as a post-processing step for improving the effectiveness of the existing entity extraction technique. These techniques utilise models trained with the web-scale corpora which makes our techniques robust and versatile. Experiments show that our techniques bring a notable improvement on efficiency and effectiveness.

\end{abstract}

\section{Introduction}
Entity extraction is widely used in text mining and natural language processing. For example, it can be used for pre-processing unstructured text: tagging and highlighting the named entities of interest. A common approach for approximate entity extraction is by comparing a substring against an entity. The approach identifies the candidate substrings from free text that match a given list of named entities. For ease of presentation, we use ``dictionary'' to refer to the list and ``entities'' to refer to the named entities. Our previous work~\cite{wen2019efficient} developped the ``2ED'' algorithm, which this paper is built on, represents a string matching approach for entity extraction. 2ED is based a distance that considers both character-level edit-distance and token-level edit-distance between a substring from the text and an entity from the dictionary.

Although 2ED reaches a high F$_1$ score with improved efficiency compared to other techniques, the limitation of 2ED is that 2ED is based on lexical evidence of the text and the dictionary, which lacks the ability to catch syntactical and semantical evidence within the text and the dictionary. To improve 2ED, we propose multiple techniques including (i) using web-scale corpora for distinguishing a typo from an intended token in the substring, (ii) estimating word similarity using word embedding, and (iii) other improvements including more advanced tokenisation. We implement our proposed techniques as a post-processing step for the 2ED algorithm. According to our evaluation, the post-processing brings 47\% improvement measured by area under the Receiver Operating Characteristic (ROC) curve, in predicting whether a matched substring represents a valid entity in the dictionary. 

%The rest of the paper consists of 7 parts. In Section~\ref{paper:rw}, we will presnt the related work in entity extraction. In Section~\ref{paper:2ed}, we discuss features and drawbacks of the previous 2ED algorithm. Section~\ref{paper:improvement} is dedicated to elaborating the improvements made for the 2ED algorithm. Section~\ref{paper:implementation} and~\ref{paper:exp} describe the implementation and validation of these improvements. Section~\ref{paper:con-fur} draws conclusion and directs future work in this topic.

\section{Related Work}
\label{paper:rw}

Another widely adopted approach to entity extraction is machine Learning such as Recurrent Neural Networks (RNN)~\cite{wei2016disease}, Convolutional Neural Networks (CNN)~\cite{chiu2016named} and SVMs~\cite{wen2014mascot}. Most of the machine learning based approaches do not require a dictionary consisting of entities of interest. In this paper, we mainly focus on the string matching approach which finds the nearest neighbour of an entity~\cite{jagadish2005idistance}. There have been various research in the area of entity extraction. Chiu et al.~\cite{chiu2016named} used an architecture combining Long Short-Term Memory (LSTM) and CNN for named entity extraction, which can utilise both token-level and character-level evidence. Wei et al.~\cite{wei2016disease} applied an RNN to entity extraction tasks in the medical domain. Besides the machine learning approaches, there are also researches focused on string matching for entity extraction, like the concept recognition system in~\cite{tseytlin2016noble} which implemented a dictionary based entity extraction tool as a component of an NLP pipeline.

Recent works in language models also inspired our work in this paper. Heafield ~\cite{heafield2011kenlm} proposed an efficient language model enabling fast queries which also uses space-efficient data structures like TRIE. The BerkeleyLM project~\cite{heafield2011kenlm} enables the storage of large n-gram language models with a fast and small data structure.
Progress in the research of word embedding has facilitated many text mining tasks. Mikolov et al.~\cite{mikolov2013linguistic} explored the performance of representations of words in a vector-space and demonstrated that these vectors captures various features and rules in the language without human intervention during the training phase.
Whitelaw et al.~\cite{whitelaw2009using} proposed an spell-checking and autocorrection method that utilises the web as a noisy corpus. The approach used shares some common features with our method in this paper as it makes use of the web as a source for training a language model. One major difference is that in their approach a machine learning based classifier was further used while we rely on a rule-based method for classification.

\section{The 2ED Algorithm}
\label{paper:2ed}

The previous 2ED algorithm proposed a novel method for estimating the distance (similarity) between a candidate substring and an entity called FuzzyED~\cite{wen2019efficient}. The novelty of FuzzyED is in that it proposed a function for measuring similarity between two strings that consist of a sequence of tokens, by taking into consideration both the character-level and the token-level edit-distances. Together with the function is a series of techniques for improving the performance by producing highly promising candidate sub-strings in an efficient way~\cite{wen2019efficient}.
We will first introduce some key features of the 2ED algorithm which are related to the post-processing work that we will discuss in this paper. We will also point out some potential weaknesses of this previous work that enlightened the improvements we will propose.

\subsection{Features of the 2ED Algorithm}

\textit{Using IDF to assign weights to tokens}:
One important idea that is exploited throughout the process of 2ED is that tokens in an entity should have discriminated weights. The algorithm proposed using (normalised) Inverted Document Frequency (IDF) as the weight for each token~\cite{wen2019efficient}, which makes use of information from the dictionary. The IDF of a token in the dictionary is a representation of its relative importance in an entity. If a token is rarely seen in the dictionary, the 2ED algorithm assumes it is more substantial in a named entity. Such tokens are called ``core'' tokens and should form an essential part of an entity~\cite{wen2019efficient}. The weights of tokens are widely used in several steps of the algorithm including the sub-string generation step, where ``core'' tokens are used as the starting point for spanning; the spanning step, for determining the point for terminating spanning of the sub-string; the shrinking step, for updating the lower bound dissimilarity; and finally the computation of FuzzyED.

\textit{Pruning and filtering}:
Since the computation of FuzzyED is relatively expensive, multiple methods for reducing the number of candidates are proposed. The algorithm uses sophisticated spanning and shrinking techniques for generating candidate substrings from the text which is proven to be more efficient than enumeration based substring generation algorithms~\cite{wen2019efficient}. In addition, some general filters utilizing information from IDF are used to further reduce the candidates for FuzzyED computation.

\textit{Calculation of FuzzyED score}:
The final step for determining whether a candidate matches an entity is to calculate similarity between the two strings by applying the FuzzyED algorithm. The formal definition is $FuzzyED(E, S)=C_D(S)+C_I(E)+C_S(E, S)$, were $E$ and $S$ denotes the entity and the substring respectively. $C_D(\cdot)$ denotes the deletion cost of removing a token; $C_I(\cdot)$ denotes the insertion cost of inserting a token; $C_S(\cdot)$ denotes the substitution cost of substituting a token in $S$ with a similar token in $E$.
 
As we can see, both character-level and token-level edit-distances are considered. The character-level edit-distance is in the substitution cost part, where the cost of substituting a token with another is related to the edit-different of the two tokens. The token-level edit-distance is calculated in a way similar to the character-level version, which is a well studied dynamic programming problem. The resulting FuzzyED score is in the range of [0,1] representing the similarity between the substring and the entity, where a score of 1 means exact match. The pairs with a FuzzyED score greater or equal to the threshold  will be added to the extracted entity list.

\textit{Parameters in 2ED}:
The 2ED algorithm features hyper-parameters that can be used to tune the thresholds for two levels of similarities~\cite{wen2019efficient}. The hyper-parameters include $\delta$ for token-level similarity threshold and $\tau$ for character-level similarity thershold. The two hyper-parameters are in domain [0,1].

\subsection{Drawbacks of the 2ED algorithm}
\textit{Applicability of IDFs as token weights}:
As described above, the 2ED algorithm uses the dictionary as the source for obtaining Inverted Document Frequency. The intuition of this approach is to assign different weights to the tokens within an entity so that they can reflect their relative importance in the entity. Such weights are finally used to calculate the FuzzyED score as in formula (1). 
However, the approach for weighing different tokens directly from the dictionary may have some potential vulnerabilities. Suppose a data analysis practitioner is interested in Australian educational institutes on Wikipedia and a dictionary specifically designed for this purpose is used. Then the IDF of tokens such as ``Australian'' and ``University'' can be much smaller than desired. In this case, IDF from a more comprehensive corpus might be a better fit. In fact, the experimental data sets used in the validation of the 2ED algorithm contain dictionaries of millions of named entities, which makes it more appropriate to use IDF as token weights.

\textit{Effectiveness of token-level edit-distance}:
The proposed function for FuzzyED calculates the cost of token-level edit by the sum of three operations: insertion, deletion and substitution. Thus the cost of transforming ``Alpha Beta'' to ``Beta Alpha'' is one deletion and one insertion, which is the sum of weights of the two words. This is because the same operations are used for token-edit and character-edit.
But as an observation from the English language, token-edit and character-edit are different. In the above example, we might have over estimated the distance between ``Alpha Beta'' and ``Beta Alpha". Consequently, more operations on the token level needs to be introduced and their costs should be studied to reflect the linguistic ``distance".
Another concern about including token-level edit-distance in the FuzzyED algorithm is its significance in real world use cases, i.e. how many matched pairs (between a substring and an entity) truly incur a token-level edit operation. We will show that this concern is valid for the corpus and dictionary we have chosen in the next section where statistics of the validation of the current 2ED algorithm will be presented.

\section{Improvement on 2ED}
\label{paper:improvement}
The effectiveness of the 2ED algorithm in terms of precision and recall was studied in previous works~\cite{wen2019efficient}. The metrics used are listed in the table below.

\begin{table}
\caption{Metrics used for evaluating 2ED}
\centering
\begin{tabular}{|c|c|c|}
\hline
notation & description    & definition \\ \hline
$tp$       & True Positive count & \# of correctly returned entities        \\ \hline
$fp$       & False Positive count & \# of wrongly returned entities         \\ \hline
$fn$       & False Negative count & \# of missed  entities \\ \hline
$p$        & Precision      & $p=\frac{tp}{tp+fp}$            \\ \hline
$r$        & Recall         & $r=\frac{tp}{tp+fn}$             \\ \hline
\end{tabular}
\end{table}

According to the study in~\cite{wen2019efficient}, 2ED reached a recall of above 96\% when token-level threshold  is set to 0.9, and above 99\% when  is set to 0.85 on some data sets. While the results above are impressive, we also explored the previous 2ED algorithm in greater detail by focusing on its performance on two kinds of edit-distances respectively. We use a lower threshold of  at 0.8 to allow us to observe as many matched pairs of substrings and entities (hereafter ``matched pairs'') as reasonably possible. The data set used at this stage is a corpus of IMDb reviews~\cite{maas2011learning} and a dictionary of movie titles obtained from the IMDb website~\cite{harper2016movielens}. The results of running 2ED algorithm on this data set are as follows: (i) the number of sub-strings matched (with $\delta=\tau=0.8$) is 39908, and (ii) the number of sub-strings approximately matched (score$<$1) is 7540. 

The table below shows a summary of the labelled matched pairs.
\begin{table}
\caption{Summary of the labelled matched pairs from 2ED}
\centering
\begin{tabular}{|c|c|c|c|}
\hline
Summary of labelled matched pairs                                      & Total                    & $fp$ & $tp$  \\ \hline
\multicolumn{1}{|l|}{number of matched pairs labelled}                                       & 200                      & 47 & 153 \\ \hline
\multicolumn{1}{|l|}{matched pairs with token-level edit-distance}                           & 42                       & 20 & 22  \\ \hline
\multicolumn{1}{|l|}{matched pairs with character-level edit-distance} & \multicolumn{1}{l|}{156} & 27 & 129 \\ \hline
\multicolumn{1}{|l|}{matched pairs with both levels of edit-distance}  & 2  & 0  & 2   \\ \hline
\end{tabular}
\end{table}

Although the number of substrings labelled is relatively small, we can still draw some qualitative conclusions:
First, the number of approximately matched substrings (score$<$1) takes a considerable proportion of the number of all matched substrings when the threshold $\delta$ is set to 0.8.
Secondly, the true positive matches takes up over 75 percent of the manually labelled sample.
Thirdly, out of the manually labelled sample, substrings with character-level edit-distance takes a dominant majority and also contributes to the biggest proportion of the true positive matches.
Lastly, substrings with both levels of edit distance takes a little proportion in the sample.
According to the analysis above, we will focus on improving the effectiveness of the previous work for extracting substrings with character-level edit-distance.

\subsection{Distinguishing a typo from an intended token}
\label{paper:typo-vs-intended}

\subsubsection{Limitation of lexical edit-distances}
Previous experimental studies of the 2ED algorithm provided some intuitions for our improvement work. For example, the following two matched pairs have a very close 2ED score but pair \#1 is an invalid match while pair \#2 is valid.

\begin{table}
\caption{Examples from 2ED}
\label{tbl:example-2ed}
\centering
\begin{tabular}{|c|c|l|l|}
\hline
\# & Substring         & Entity            & 2ED      \\ \hline
1  & about the premise & about the promise & 0.844754 \\ \hline
2  & code of honor     & code of honour    & 0.862025 \\ \hline
\end{tabular}
\end{table}

To be more general, 2ED measures the similarity of two tokens by lexical edit-distance. In example \#1, the difference between this matched pair is within the word pair ``premise'' and ``promise". 2ED measures the distance between the word pair by number of character-level operations including insertion, deletion and substitution. Thus, the difference is represented by a substitution operation that turns the letter ``e'' in word ``premise'' into the letter ``o'' in word ``promise". Similarly in example \#2, the distance is represented by an insertion operation that turns the word ``honor'' into ``honour".
However, the validity of the two matched pairs is not represented by the lexical edit-distance within these pairs. The pair in example \#2 is valid because token ``honor'' is an variation of token ``honour'' in English; while the pair in example \#1 is not valid because ``premise'' and ``promise'' are two different words that share little similarity in their grammatical position and semantical meaning.
In fact, the lexical edit-distance between two tokens is more applicable as a representation of their similarity (distance) when one of them is a mis-spelled version (typo) of the other. It is common that words that look similar may or may not have close meanings and grammatical position (e.g. part-of-speech). Thus, some criteria for distinguishing a typo from an intended token should be introduced to help us judge whether it is appropriate to apply the FuzzyED which is based on lexical edit-distance.

\subsection{Using language models}
\label{paper:lang-model}

We propose the following conditions for identifying a typo based on the above analysis: suppose the substring contains a token ts that has a close lexical edit-distance with the corresponding token te in the entity. We assume ts is not a typo if and only if (i) ts is a valid word in the language and (ii) ts fits in the context in the substring.
The next step is to model these two conditions in a feasible way. In fact, the conditions (i) and (ii) above can both be judged with a corpus of its language, where the validity of a single token can be measured by its frequency in the corpus and the validity of its context by the frequency of word phrases or (token-level) n-grams. Since single tokens are just (token-level) 1-grams, the above conditions can be simplified as judging whether the n-grams generated around ts are valid n-grams in the language, where n is in range $[1,k]$ and $k \geq 2$.
N-grams and language models
The application of (token-level) n-grams in NLP tasks is versatile, one of them being a statistical language model. A language model can help us tell (i) how likely a given n-gram will appear in a language or (ii) the conditional probability that an n-gram is followed by a certain word. According to the analysis above, we are using statistics of the n-grams in a language without domain-specific knowledge or linguistic rules. Thus it is appropriate to use a statistical language model trained from a large corpus. 

\textbf{Utilising web-scale corpora}:
Since our task is to specifically use the language model as a comprehensive corpus of a language, it is critical to find a source where we can obtain large scale n-grams. In 2012, Lin et al. [7, 8] published the second version of the ``Google Books Ngram Corpus”, where frequencies of n-grams in the Google Books collection are collected with historical statistics. The corpus ``reflects 6\% of all books ever published"~\cite{lin2012syntactic}. Considering the scale and coverage of the corpus, we find it a great source for our purpose. 
Training from such a large corpus consisting of hundreds of gigabytes of data is not trivial. Even if we leave out all historical information and use n-gram frequencies collectively, maintaining a map of n-grams and their frequencies is still a memory consuming task to be executed on a single machine. Besides memory consumption of the training process, the complexity of the model is another concern.
Based on the above complexity analysis, we adopt a relatively simple model in this paper, the BerkeleyLM~\cite{pauls2011faster}, which features a trained n-gram based model using the Google Books Ngram Corpus. The model provides interfaces for querying (i) the (conditional) log-probability of an n-gram and (ii) the raw count of an n-gram in the Google Books Ngram Corpus.
The trained model uses stupid back-off for estimating the (conditional) probability of an n-gram as follows $P(w_i|w_1, w_2, ..., w_{i-1}) = \frac{count(w_1, w_2, ..., w_i)}{count(w_1, w2, ..., w_{i-1})}$.

\subsection{Estimating word similarity}
\label{paper:est-word-sim}
As discussed in Section~\ref{paper:typo-vs-intended}, we are interested in the case where an unmatched token in the substring is not a typo and thus the applicability of the previous 2ED algorithm needs to be carefully reviewed. Here, the method we propose is to use word embedding for measuring the distance between two tokens. 
From the experimental results in Section~\ref{paper:typo-vs-intended}, an observation is that the unmatched tokens in a matched pair can belong to various cases. For example, (i) ts and te may be variations of each other (e.g. ``honour'' and ``honour"); (ii) ts may be the plural form of te or vice versa (e.g. ``survivors'' and ``survivor"). A word embedding will help us capture the ``distance'' between the unmatched tokens.
The word embedding adopted in this paper is Google's word2vec~\cite{bengio2003neural} which represents words in a corpus by vectors of floats. According to Mikolov et al.~\cite{mikolov2013linguistic,mikolov2013distributed,mikolov2013efficient}, the trained model can capture syntactic and semantic information of words in a language. 

\subsection{Other improvements}
\label{paper:other-improvement}
Besides the introduction of language models and word embedding to replace some functionality of the FuzzyED distance metric, we have also made other minor improvements.
The previous 2ED algorithm does not separate the period from the last word in a sentence during its tokenisation phrase. Since many approximately matched pairs are exact matches if we strip the period (dot) from the last token in the substring, we make this operation an optional feature in the implemented post-processing algorithm.

\section{Implementation}
\label{paper:implementation}
The improved methods described in Section~\ref{paper:improvement} is implemented in a pipeline. All improvements are applied as post-processing steps to filter, examine and (possibly) re-score matched pairs selected by previous 2ED algorithm. We will walk through the pipeline step by step in the rest of this section.

\subsection{Obtain candidate pairs}
\label{paper:candiate-pairs}

The first step is finding matched pairs with character-level edit-distance as candidates for future re-scoring work. We apply the previous 2ED algorithm to our corpus against a dictionary with token-level similarity threshold $\delta=0.8$ and character-level similarity threshold $\tau=0.8$.  was chosen according to the parameter optimization work in~\cite{wen2019efficient};  was chosen for error tolerance with the previous algorithm.
After obtaining the list of matched pairs, we filter out exact matches (i.e. 2ED score $= 1$) because they are not the part of the result we are trying to improve. For the rest of the matched pairs, we apply the following steps for each pair.

\subsection{Rescore candiadte pairs}
\label{paper:rescore}

\textbf{Filter out pairs with tokenisation problems}:
An approximately matched pairs with tokenisation problems as described in Section~\ref{paper:lang-model} is not processed into the next step. Rather, we simply strip the ending period from the substring and assume it an exact match. The stripping step is operational and improvement from this operation is separately analysed as in Section~\ref{paper:exp}.

\textbf{Generate $n$-gram to check validity}:
With an approximately matched pair, we compare each token pair in the corresponding position of the substring and the entity to identify whether there is (only) a character-level edit-distance between this pair. In this process, we have also obtained the position of the ts and te as per notation in Section~\ref{paper:typo-vs-intended} if the token pair does exist.
We then generate (token-level) n-gram pairs surrounding ts and te in the substring and the entity term respectively. We first generate 3-grams. If the substring is too short that it contains less than 3 tokens. We use the substring and the entity as a whole, i.e. 2-grams or 1-gram, for a pair.
The next step is to check the validity of these n-grams in a language model to help us distinguish a typo from an intended word according to the conditions described in Section~\ref{paper:typo-vs-intended}. We use a tolerant criteria for this validity check, where the unmatched token in the substring ts is considered an intended word as long as any n-gram pair from ts and te are both valid or both invalid in the language model.
For each (token-level) n-gram, we use two thresholds accounting for the log-probability and raw count in the language model respectively to help check its validity. The thresholds are set according to empirical observations. The threshold for log-probability is -10.8 and the threshold for raw count is 0.

\textbf{Apply cosine similarity to rescore}:
When an unmatched token is identified as an intended word in the last step, we use the trained word2vec embedding to calculate the similarity between the token pair ts and te. For the examples in Section~\ref{paper:typo-vs-intended}, the cosine similarity are shown in the table below. As we can see, these scores represents the distance between the token pairs in English: the similarity between ``honor'' and ``honour'' is significantly higher than that between ``premise'' and ``promise'' since the former consists of variations of the same word while the latter consists of two distinct words.
\begin{table}
\caption{Cosine similarity for examples in Table~\ref{tbl:example-2ed}}
\centering
\begin{tabular}{|c|c|l|l|}
\hline
\# & Substring         & Entity            & Cosine similarity \\ \hline
1  & about the premise & about the promise & 0.245628          \\ \hline
2  & code of honor     & code of honour    & 0.637478          \\ \hline
\end{tabular}
\end{table}

The cosine similarity between ts and te is normalised using the following formula. This formula is chosen according to empirical observation of the distribution of the cosine similarity (denoted by cos in the formula). It guarantees the following features: (i) normalised edit-distance is 0 when cos is 1, i.e. the edit-distance is 0 for two identical words; (ii) normalised edit-distance is 1 whencos is 0 (although cos is within range [-1,1], we observe that most empirical results sits in [0,1]); (iii) normalised edit-distance punishes low cos scores using an exponential formula; (iv) base is a tunable parameter affecting the curve of the normalization function: $ED_{norm} = \frac{base^{1 - cos} - 1}{base - 1}$. A final score is applied to the post-processed pair of substring and entity using the following formula. We take the length of the entity as a normalizing parameter. This approach is similar to the normalization in FuzzyED when we assign a uniform weight to the tokens in the entity: $Rescore = 1 - \frac{ED_{norm}}{length(entity)}$.

\section{Experimental studies}
\label{paper:exp}

\textit{Evaluation setup}:
The validation was performed on the NeCTAR research cloud~\cite{corbet1984honeybees} using a 12-core computing instance with 48 GB of RAM. The test data set was obtained from a public corpus of Amazon reviews~\cite{mcauley2013hidden}. The corpus consists of (i) millions of reviews on the Amazon website, further divided into subsets by product category; (ii) metadata of the products available on the Amazon website.
From the review data set, we selected the books subset which contains more than 8 million reviews and sampled 1000 reviews as the text for the task. From the metadata which contains information about 9.4 billion product items, we extracted only the titles of these items and use the result as the dictionary for the task. Due to missing fields in the metadata, the dictionary consists of 7.99 million product titles.

The corpus for training word2vec embeddings is obtained from various resources on the web using the script provided in the toolset on~\cite{bengio2003neural}. 
The resulting training set consists of 6.1 billion tokens. After two runs of the word2phrase pre-processing~\cite{bengio2003neural}, we train the word2vec embedding with the Continuous Bag of Word (CBOW) method and a vector size of 300 dimensions on the data set. Another training method skipgram was also attempted but achieved lower precision using the provided validation tool in~\cite{bengio2003neural}; and it was not used in future steps.
The training process takes less than a day to finish on the cloud instance.

We use (i) the distribution of the scores of re-visited pairs for positive pairs and negative pairs respectively, and (ii) the Receiver Operating Characteristic (ROC) curve to validate the effectiveness of the post-processing. We manually labelled 113 pairs from the re-scored set, which comprises over 10\% of its size and use that labelled data to evaluate the result. 

\begin{figure}
%\vspace{-10pt}
\centering
\subfloat[\vspace{50pt} Label = Y]{
\includegraphics[width=0.49\textwidth, height=1.4in]{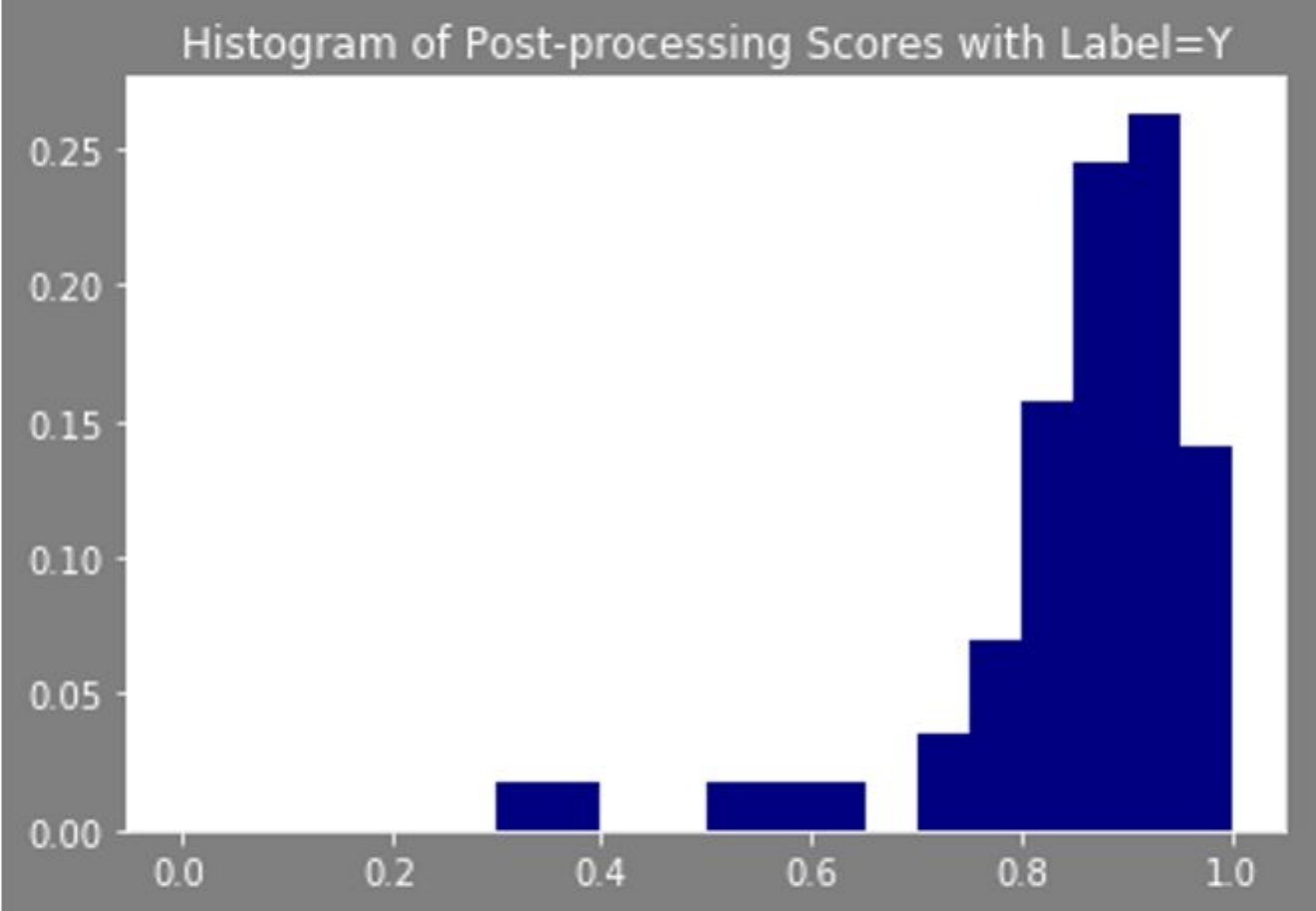}
}
\subfloat[\vspace{50pt} Label = N]{
  \includegraphics[width=0.49\textwidth, height=1.4in]{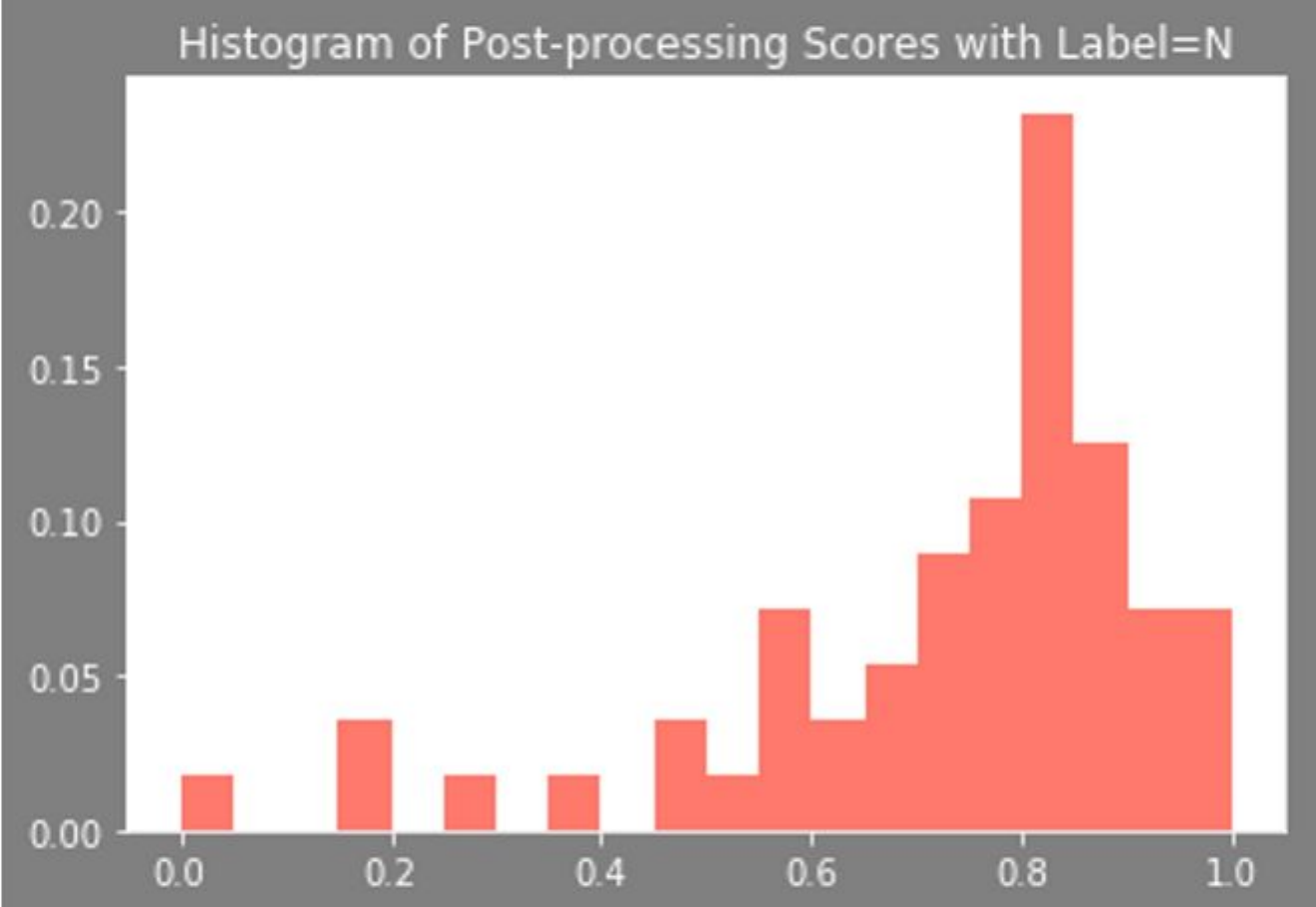}
}
\vspace{-55pt}
\caption{Histogram of post-processing scores}
\label{fig:hist-post-proc}
\end{figure}

\textit{Effectiveness}:
Figure~\ref{fig:hist-post-proc} shows the distribution of the re-visited scores separated by their labels, where label Y means the matched pair is valid (according to human evaluation) and label N means the pair is not. As we can see, by applying the post-processing, the two groups have their scores distributed in two clusters in distinct centroids.

The two ROC curves below compare the performance of using 2ED score and post-processed score to predict validity of extracted substrings. As we can see, the post-processed score achieves a higher true positive rate without sacrificing the false positive rate, while 2ED score performs like random guess in evaluation of extracted substrings. Overall the post-processed score achieved an area under curve (AUC) of 0.72, and outperforms 2ED score by 47\%.

\begin{figure}
%\vspace{-10pt}
\centering
\subfloat[\vspace{50pt} Label = Y]{
\includegraphics[width=0.48\textwidth, height=1.4in]{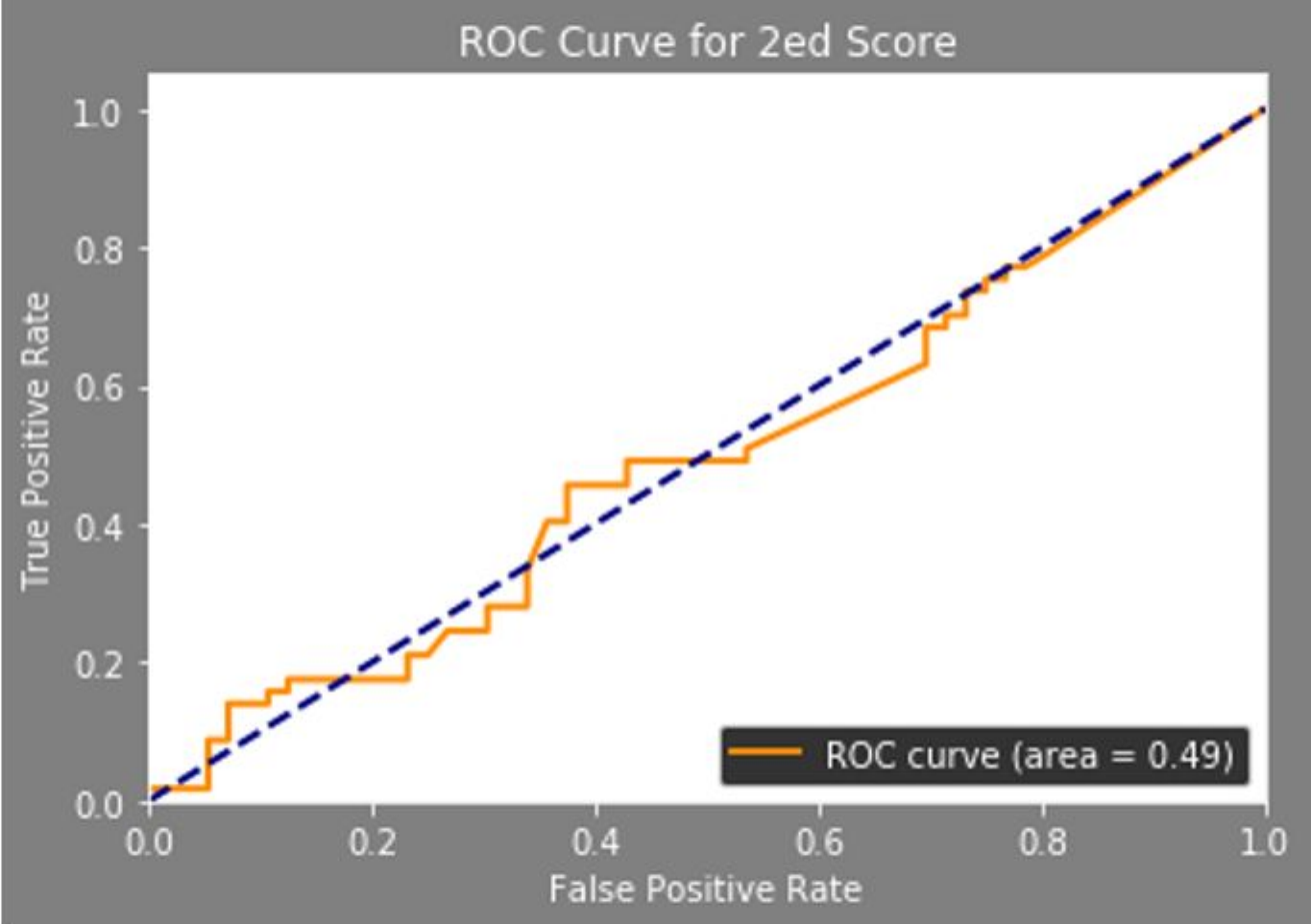}
}
\subfloat[\vspace{50pt} Label = N]{
  \includegraphics[width=0.48\textwidth, height=1.4in]{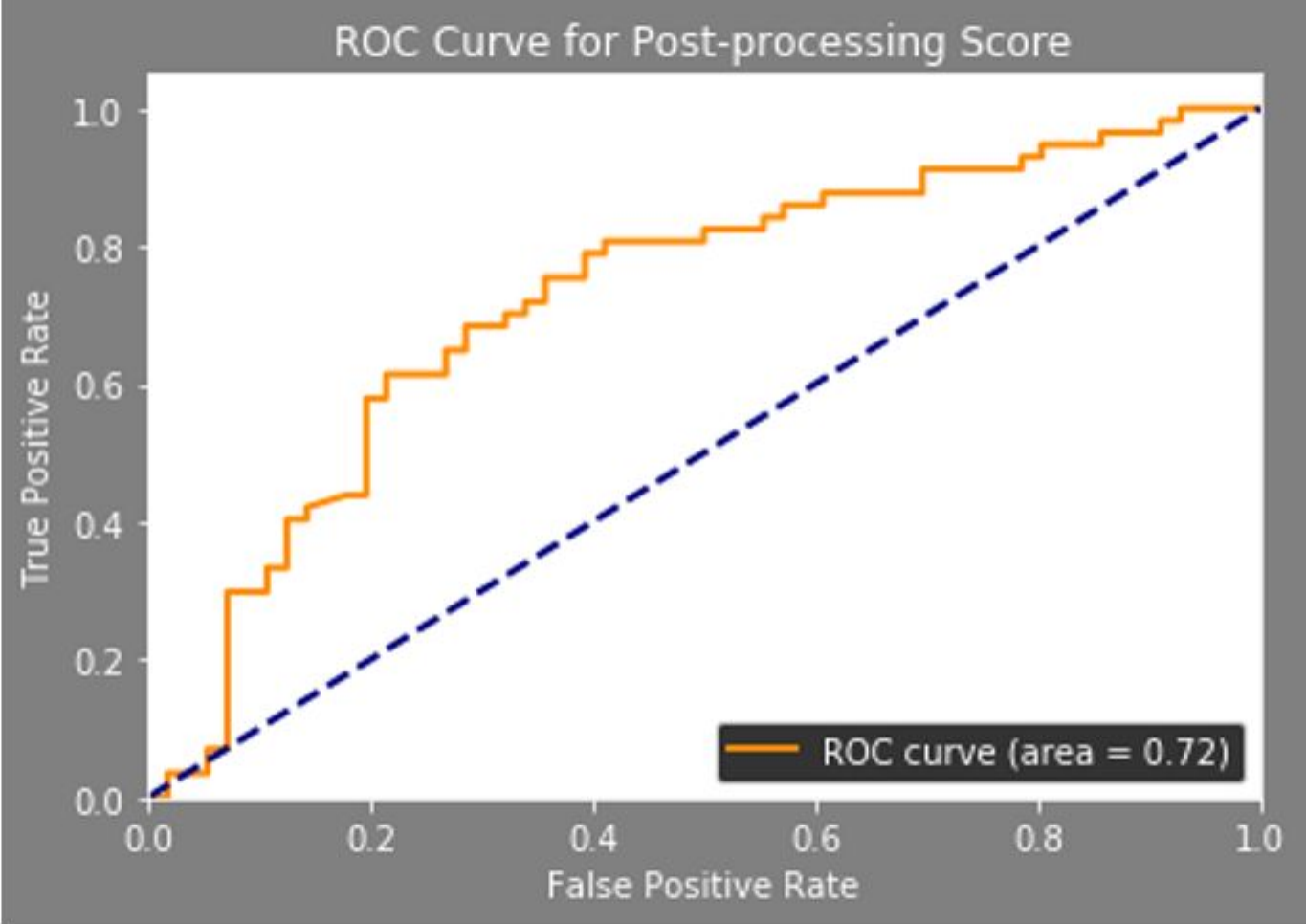}
}
\vspace{-55pt}
\caption{ROC curves for 2ED score and post-processing score}
\label{fig:roc}
\end{figure}

\textit{Efficiency}
The efficiency of the algorithm is not the major concern in this paper. We evaluate the performance of the post-processing algorithm by the time taken to complete the task on the data set described earlier in this section. 
The task typically finishes within 10 minutes, depending on the status of the cloud instance. Majority of the time taken is on loading trained models into memory, so the performance should also depend on the physical RAM available on the evaluation instance. After loading the models, the application finishes processing over 2000 items in less than 5 seconds. 
Therefore, this algorithm should be suitable as a post-processing step for the previous 2ED algorithm in terms of its performance.

\section{Conclusion and Future Work}
\label{paper:con-fur}
In this paper, we proposed several improvements to our previous entity extraction algorithm called ``2ED''. Our proposed improvements include language models for typo detection, word embedding to measure word distances to capture semantic features, and more advanced tokenization. We have implemented the proposed techniques as a post-processing step on top of 2ED. Our proposed techniques bring significant improvement to 2ED. The improvement mainly lies in the introduction of web-scale corpora used for training relatively comprehensive and versatile models. This finding shows that more information from the web-scale corpora can facilitate entity extraction.

Some improvements and extensions to this work can be made to further generalise its applicability, boost its performance and make better use of the web-scale corpus. First, it is possible to combine evidence from postags. The new version of Google Books Ngram Corpus features part-of-speech tag (i.e., postag) information. Such labels can be further utilised for measuring the distance between a token pair in addition to the $n$-gram used in the current implementation. Furthermore, beyond the post-processing approach, postags can facilitate the candidate substring generation process in 2ED. Second, it is promising to tokenise text with punctuations. 2ED uses a tokenisation method which does not separate the period of a sentence with its ending word. One major concern of stripping periods from the ending words in the previous implementation is it is hard to distinguish a ``true'' punctuation which ends a sentence or clause from an ending dot of an abbreviation lexically. With the language models introduced in this paper, potentially we are able to find effective ways to improve the tokenisation using linguistic evidence from these models. Third, learning parameters for 2ED is also helpful for users. The current implementation uses empirical settings of parameters for judging $n$-gram validity and for normalising the cosine similarity of a token pair. These parameters can be learned from labelled data. Finally, our approach extracts entities in English, and our approach can be extended to other languages.

\bibliographystyle{abbrv}
\bibliography{FuzzyED}

\end{document}